\batchmode
\makeatletter
\def\input@path{{/home/yecheng/ISCAS2019/}}
\makeatother
\documentclass[twocolumn,english]{IEEEtran}
\usepackage[T1]{fontenc}
\usepackage[latin9]{inputenc}
\usepackage{array}
\usepackage{amsmath}
\usepackage{amssymb}
\usepackage{graphicx}

\makeatletter

\providecommand{\tabularnewline}{\\}

\usepackage{cite}

\usepackage{babel}
\usepackage{subfig}
\usepackage{caption}

\makeatother

\usepackage{babel}
\begin{document}
\title{\pagenumbering{gobble}Road Segmentation Using CNN and Distributed
LSTM}
\author{Yecheng Lyu, Lin Bai and Xinming Huang\\
 Department of Electrical and Computer Engineering\\
Worcester Polytechnic Institute\\
Worcester, MA 01609, USA\\
\{ylyu,lbai2,xhuang\}@wpi.edu}
\maketitle
\begin{abstract}
In automated driving systems (ADS) and advanced driver-assistance
systems (ADAS), an efficient road segmentation is necessary to perceive
the drivable region and build an occupancy map for path planning.
The existing algorithms implement gigantic convolutional neural networks
(CNNs) that are computationally expensive and time consuming. In this
paper, we introduced distributed LSTM, a neural network widely used
in audio and video processing, to process rows and columns in images
and feature maps. We then propose a new network combining the convolutional
and distributed LSTM layers to solve the road segmentation problem.
In the end, the network is trained and tested in KITTI road benchmark.
The result shows that the combined structure enhances the feature
extraction and processing but takes less processing time than pure
CNN structure.
\end{abstract}


\section{Introduction}

In recent years, growing research interest is witnessed in automated
driving systems (ADS) and advanced driver-assistance systems (ADAS).
As one of the essential modules, road segmentation perceives the surroundings,
detects the drivable region and builds an occupancy map \cite{hillel2014RoadDetection}\cite{alahi2016LSTM_Human_Traj}\cite{lyu2018ISCAS}\cite{lyu2018chipnet}.
A drivable region is a connected road surface area that is not occupied
by any vehicles, pedestrians, cyclists or other obstacles. In the
ADS workflow, road segmentation contributes to other perception modules
and generates an occupancy map for planning modules. Therefore, an
accurate and efficient road segmentation is necessary.

Camera-based road segmentation has been investigated for decades since
cameras generate high-resolution frames frequently and they are cost
effective. Traditional computer vision algorithms employed manually
defined features such as edges \cite{yoo2013gradient} and histogram
\cite{yoo2013IlluminationRobustLaneDetection} for road segmentation.
Those features, however, worked on limited situations and were difficult
to extend to new scenarios \cite{hillel2014RoadDetection}.

Convolutional neural network (CNN) based algorithms attracted research
interest in recent years. By implementing massive convolutional kernels
to a deep neural network, CNNs are capable to handle various driving
scenarios. Existing CNN based road segmentation algorithms such as
FCN \cite{long2015fully}, SegNet \cite{badrinarayanan2015Segnet},
StixelNet \cite{levi2015stixelnet}, Up-conv-Poly \cite{oliveira2016UpConv}
and MAP \cite{laddha2016map} generated a precise drivable region
but required large computational. Table \ref{Table: Evaluation result}
presents their performance on KITTI road benchmark as well as their
parameter counts, floating-point operations and running time for each
frame processing. Recent research proposed several efficient networks
and network structures such as MobileNet \cite{Howard2017MobileNets_Google}
and Xception \cite{DBLP:journals/corr/Chollet16a}. However, they
were still too large to work on embedded systems. In our previous
work \cite{zhou2018RoadNet} we explored the use of CNN stack to process
the camera data and implemented on a VLSI die but experienced high
memory usage.

Long-Short Term Memory (LSTM) is a kind of recurrent neural networks
(RNNs) that are often used to process streaming data such as an audio
signal and video sequences. By introducing memory cells and gates,
LSTM units are capable to extract context features in a long sequence
of inputs. Recently, distributed LSTMs are introduced to share the
LSTM kernel weights in multi-sequence processing. However, LSTM and
distributed LSTM algorithms only focus on time-domain processing.

In this paper, we introduce the distributed LSTM to work on spatial
domain and process row sequences on camera frames and corresponding
feature maps. This is one of the first efforts to LSTM on spatial
sequence processing. We also propose a deep neural network that combines
the convolutional layer and distributed LSTM layers on the spatial
domain to solve road segmentation tasks. The proposed network is trained
and tested on the KITTI road benchmark \cite{fritsch2013KITTI_road}.
It is shown that the proposed method achieves comparable accuracy
to the existing solutions but takes fewer calculations and less processing
time. The result shows that LSTM structures significantly enhance
the context feature extraction in a large feature map. The rest of
the paper is organized as follows: Section \ref{sec:Comparison of CNN and distributed LSTM}compares
the scheme of the convolutional layer and the distributed LSTM layer
on feature map processing. Section \ref{sec:RoadNet-v2} introduces
the proposed network that combines CNN and distributed LSTM. In Section
\ref{sec:Network implementation =000026 training}, we evaluate the
proposed network in the KITTI road benchmark. Section \ref{sec:Conclusion}
concludes the paper.

\section{Comparison of convolutional layer and distributed LSTM layer\label{sec:Comparison of CNN and distributed LSTM}}

In this section, we compare the scheme, sensing area and computational
complexity of CNN and distributed LSTM structure in a feature map
processing and try to show their advantages and disadvantages in feature
map processing.

\begin{figure}
\begin{centering}
\begin{tabular}{cc}
\multicolumn{2}{c}{\includegraphics[width=0.9\columnwidth]{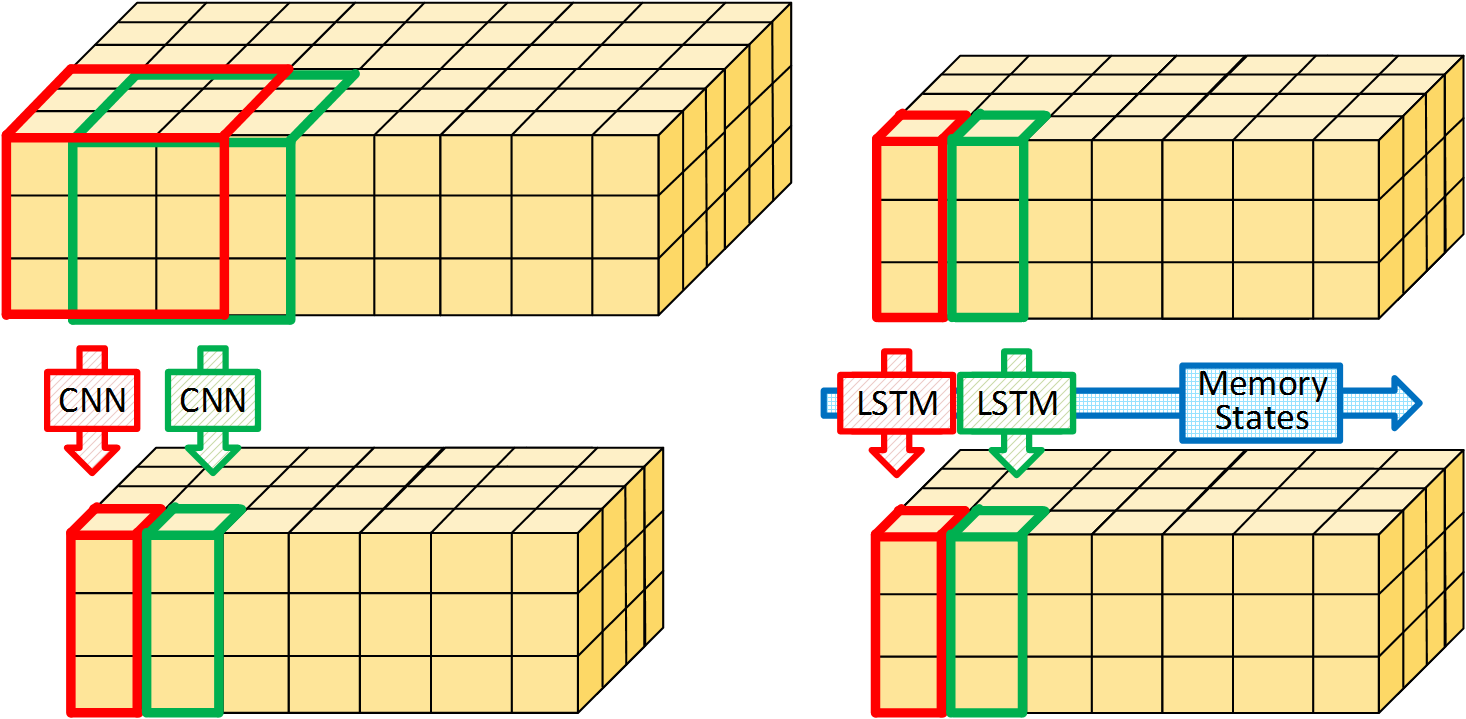}}\tabularnewline
(a) Convolutional Layer &
(b) Distributed LSTM layer\tabularnewline
\end{tabular}
\par\end{centering}
\caption{Comparison of convolutional layer and distributed LSTM layer schemes\label{Fig: Scheme of Convolutional Neural Network}}
\end{figure}

Suppose we have a feature map that is formatted as a $w_{1}\times h_{1}\times d_{1}$
tensor. A convolutional kernel of size $w_{k}\times h_{k}\times d_{x}$,
stride $(s_{w},s_{h})$ and padding $(p_{w},p_{h})$ is designed to
process the feature map. The scheme is described in Figure \ref{Fig: Scheme of Convolutional Neural Network}(a).
The kernel initially convolutes the first patch in red and generates
a $1\times1\times d_{k}$ vector. Then the kernel strides certain
pixels to the right and convolutes the next patch. After all available
patches are convoluted, a $w_{2}\times h_{2}\times d_{2}$ output
tensor is generated. (1) - (3) presents the relationship between the
size of output tensor, input tensor and kernel settings. In each step
there are $w_{k}\times h_{k}\times d_{1}$ inputs contribute to a
$1\times1\times d_{k}$ output vector. Floating-point operations for
this layer is $w_{2}\times h_{2}\times d_{1}\times w_{k}\times h_{k}\times d_{k}\times2$.

\begin{equation}
w_{2}=\left[\frac{w_{1}+2p_{w}}{s_{w}}\right]
\end{equation}

\begin{equation}
h_{2}=\left[\frac{h_{1}+2p_{h}}{s_{h}}\right]
\end{equation}

\begin{equation}
d_{2}=d_{k}
\end{equation}

On the other hand, suppose we have a feature map of the same size,
and we now have $h_{1}$ distributed LSTM units with $d_{l}$ memory
cells. The units share the same weights and bias but work on different
rows of the feature map. In addition, we configure the LSTM units
to return an output vector in each step. The scheme is illustrated
in Figure \ref{Fig: Scheme of Convolutional Neural Network}(b). In
each step, each distributed LSTM unit inputs a $1\times1\times d_{1}$
vector, updates its memory states through gate functions and output
a $1\times1\times d_{l}$ vector. Therefore, the distributed LSTM
layer generates a $w_{1}\times h_{1}\times d_{l}$ output tensor.
In the $i^{th}$ step, there are $i\times d_{1}$ inputs contribute
to a $1\times1\times d_{l}$ output tensor. Floating-point operations
for this layer is $w_{1}\times h_{1}\times8\times d_{1}\times d_{l}$.

Comparing the convolutional layer and distributed LSTM layer, we have
the following conclusions:

(1) If convolutional layer and distributed LSTM layer have the same
output size, the ratio of the floating-point operations is $w_{k}\times h_{k}:4$.
Considering that $w_{k}\geqslant3$ and $h_{k}\geqslant3$ , a distributed
LSTM layer save at least 56\% floating-point operations than a convolutional
layer.

(2) If convolutional layer and distribute LSTM layer have the same
output size, the ratio of the average number of contributing input
cells is $9:w_{1}/2$. Considering that $w_{1}$ is hundreds to thousands
in a feature map, a distributed LSTM layer has a much larger sensing
area than convolutional layers.

(3) Each kernel in a convolutional layer has a sensing area covering
multiple rows but each unit in a distributed LSTM layer works on a
single row. Therefore, a convolutional layer has a better feature
extraction vertically.

(4) A convolutional layer can generate a smaller feature map by setting
the $s_{w}$ and $s_{h}$ greater than one, but a distributed LSTM
layer cannot. Therefore, in a deep neural network, a convolutional
layer is useful to narrow the feature map to enlarge the equivalent
sensing area for the following layers and save their floating-point
operations.

(5) We can combine the convolutional layer and distributed LSTM layer
layers in a deep neural network, and the new network should take fewer
calculations but has an enhanced ability to extract context features
in horizontal.

\section{Proposed Network \label{sec:RoadNet-v2}}

\begin{figure}
\begin{centering}
\includegraphics[width=0.9\columnwidth]{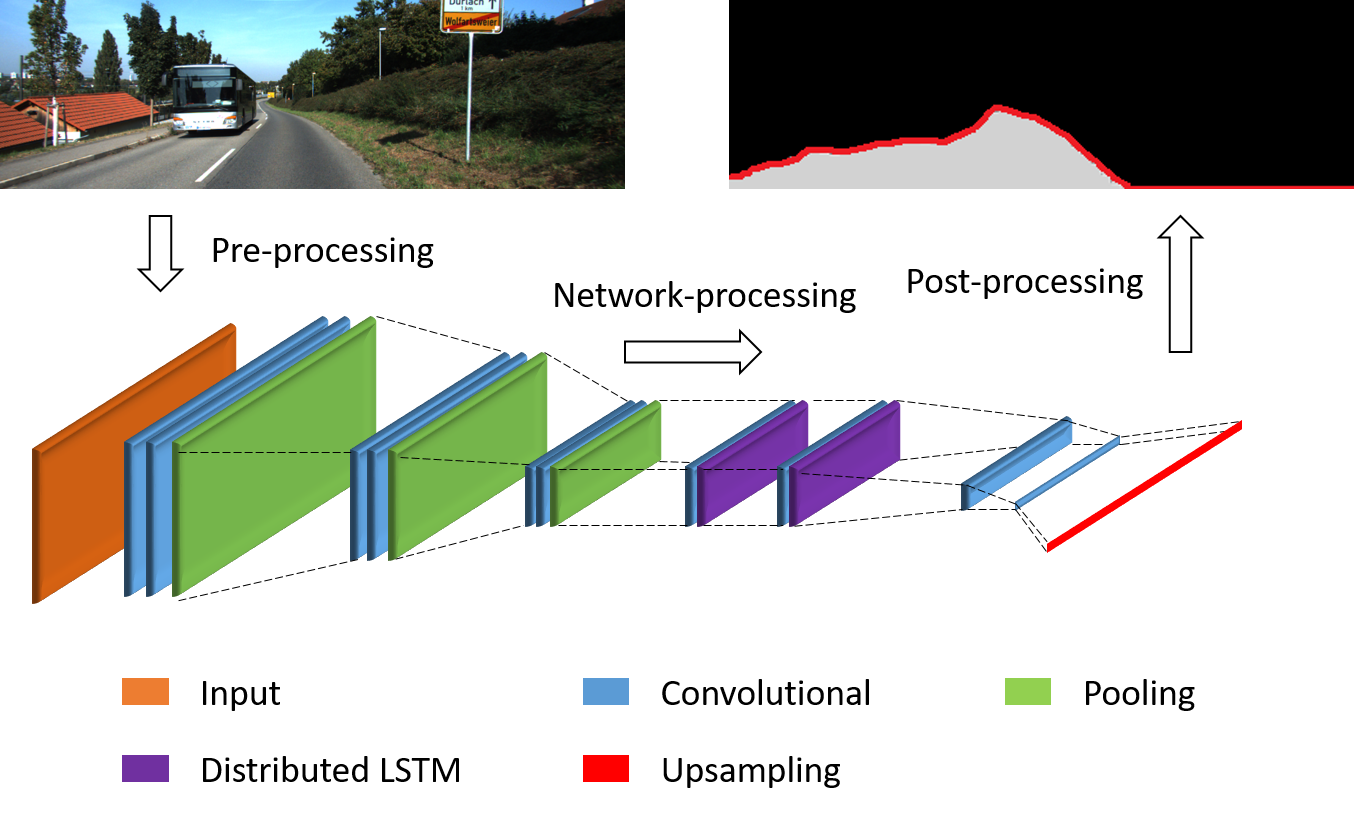}
\par\end{centering}
\caption{Diagram of the proposed network\label{fig: Network-Diagram}}

\end{figure}

This section introduces the proposed neural network. The network targets
to solve a sequential regression problem whose input is a multi-layer
feature and whose output is a row vector that has the same width of
the input. The network contains three (1) a CNN based local feature
encoder, (2) a CNN and LSTM based feature processor, and (3) a CNN
based output decoder. Figure \ref{fig: Network-Diagram} presents
an overview of network architecture.

\subsection{Local feature encoder}

The local feature encoder is a group of convolutional layers designed
to extract local features and narrow the feature map size. Commonly
a CNN based encoder cascades several convolutional layers with small
kernels into blocks because those blocks have fewer parameters to
train but work better in non-linear feature extraction. Those blocks
are obvious in FCN \cite{long2015fully}, SegNet \cite{badrinarayanan2015Segnet},
and StixelNet \cite{levi2015stixelnet}. In our work, we cascade six
convolutional layers with a $3\times3\times64$ kernel and group them
two by two. The first layer in each group is configured to $stride=(2,2)$
and the second layer is configured to $stride=(1,1)$. Therefore,
in each group the size of output tensor is half the size of its input
tensor in both in both axes. Eventually an input tensor of size $w_{1}\times h_{1}\times d_{1}$
will result in a $\frac{w_{1}}{8}\times\frac{h_{1}}{8}\times64$ output
tensor.

\subsection{Feature Processor}

The feature processor is designed to extract context features over
the entire feature map. It is built by several blocks of convolutional
layers and distributed LSTM layers. Each block includes a convolutional
layer of $3\times3\times64$ kernel and a distributed LSTM layer of
$64$ memory states. In the feature processor blocks, there is no
pooling layer added so that the input and output tensor have the same
size. 

\subsection{Output decoder}

The output decoder is designed to decode the output tensor of the
feature processor to a final row vector. The decoder upsamples the
feature map in horizontal but further narrow the feature map in vertical.
Since our decoder concentrates on the upper boundary of the drivable
region that connects to the bottom of image frame, it has fewer layers,
parameters and calculations than the decoder in SegNet. The decoder
includes several convolutional layers, and an upsampling layer. The
output size is $w_{1}\times1\times1$.

\section{Network implementation, training and testing \label{sec:Network implementation =000026 training}}

In this section, we evaluate the performance of the proposed network
using KITTI road benchmark. We first introduce the benchmark and then
present our solution using the proposed network. We also present the
training scheme for the training set and test result from the testing
set.

\subsection{KITTI road benchmark}

KITTI road benchmark is a widely used benchmark for drivable region
segmentation. It contains $290$ training samples and $289$ testing
samples recorded in a real driving scenario. The size of camera frames
in the benchmark is $1242\times375$ or $1224\times370$. The drivable
region is manually labeled in a binary map of the same size. In the
KITTI road benchmark, F1-score (F1) and average precision (AP) are
the main metrics to evaluate the accuracy of road segmentation solutions.
Other metrics are precision (PRE), recall (REC), false positive rate
(FPR), false negative rate (FNR) and running time. The metrics are
calculated as in (4) - (9) where TP, TN, FP, FN denote true positive,
true negative, false positive, and false negative.

\begin{equation}
PRE=\frac{TP}{TP+FP}
\end{equation}

\begin{equation}
REC=\frac{TP}{TP+FN}
\end{equation}

\begin{equation}
FPR=\frac{FP}{TP+FP}
\end{equation}

\begin{equation}
FNR=\frac{FN}{TP+FN}
\end{equation}

\begin{equation}
F1=\frac{2\cdot PRE\cdot REC}{PRE+REC}
\end{equation}

\begin{equation}
AP=\frac{TP+TN}{TP+FP+TN+FN}
\end{equation}

\subsection{Scheme of road segmentation }

\subsubsection{Pre-processing}

In pre-processing, we apply a pyramid approach that generates two
input tensors from each camera frame. The first input tensor focus
on near-range segmentation and the second one focus on far-range segmentation.
For the first one, the camera frame is scaled to size $600\times160$.
For the second input tensor, however, the camera frame is cropped
to size $600\times160$ in the center. Those two images are then fed
into the tensors together with their horizontal and vertical indexes
in their own image coordinate. The pyramid scheme narrows the difference
between local features in the near range and far range. It also enlarges
the training set for better parameter tuning. After pre-processing,
those two input tensors of size $600\times160\times5$ are processed
by two network instances in parallel.

\subsubsection{Network implementation}

The proposed network is implemented in the Keras platform with TensorFlow
backend as described in Section \ref{sec:RoadNet-v2}. The details
of each layer are presented in Table \ref{Table:Detailed-Network-Blocks}.
The network has 348,801 parameters and takes 3.45 billion floating-point
operations to process each input tensor. Table \ref{Table: Evaluation result}
compares the proposed network with related works on accuracy, network
parameters, floating-point operations and running time for each camera
frame. The proposed network has only 24\% parameters and takes 1.36\%
floating-point operations when comparing to SegNet.

\begin{table}
\caption{Detailed Network Blocks\label{Table:Detailed-Network-Blocks}}

\centering{}%
\begin{tabular}{|>{\centering}m{1.4cm}|>{\centering}m{1.7cm}|>{\centering}m{1.6cm}|>{\centering}m{1.7cm}|}
\hline 
Layer &
Kernel

$(w_{1}\times h_{1}\times d_{1})$ &
Input

$(w\times h\times d)$ &
Output

$(w_{2}\times h_{2}\times d_{1})$\tabularnewline
\hline 
Input &
$-$ &
$600\times160\times5$ &
$-$\tabularnewline
\hline 
Conv1

s=2 &
$3\times3\times64$ &
$600\times160\times5$ &
$600\times160\times64$\tabularnewline
\hline 
Conv2 &
$3\times3\times64$ &
$600\times160\times64$ &
$600\times160\times64$\tabularnewline
\hline 
Conv3

s=2 &
$3\times3\times64$ &
$300\times80\times64$ &
$300\times80\times64$\tabularnewline
\hline 
Conv4 &
$3\times3\times64$ &
$300\times80\times64$ &
$300\times80\times64$\tabularnewline
\hline 
Conv5

s=2 &
$3\times3\times64$ &
$150\times40\times64$ &
$150\times40\times64$\tabularnewline
\hline 
Conv6 &
$3\times3\times64$ &
$150\times40\times64$ &
$150\times40\times64$\tabularnewline
\hline 
Conv7 &
$3\times3\times64$ &
$75\times20\times64$ &
$75\times20\times64$\tabularnewline
\hline 
D-LSTM 1 &
$64$ &
$75\times20\times64$ &
$75\times20\times64$\tabularnewline
\hline 
Conv8 &
$3\times3\times64$ &
$75\times20\times64$ &
$75\times20\times64$\tabularnewline
\hline 
D-LSTM 2 &
$64$ &
$75\times20\times64$ &
$75\times20\times64$\tabularnewline
\hline 
Conv9 &
$1\times5\times64$ &
$75\times20\times64$ &
$75\times4\times64$\tabularnewline
\hline 
Conv10 &
$1\times4\times1$ &
$75\times4\times64$ &
$75\times1\times$1\tabularnewline
\hline 
Up-

sample &
$8\times1\times1$ &
$75\times1\times1$ &
$600\times1\times1$\tabularnewline
\hline 
Output  &
$-$ &
$-$ &
$600\times1\times1$\tabularnewline
\hline 
\end{tabular}
\end{table}

\subsubsection{post-processing}

\begin{figure}
\includegraphics[width=0.95\columnwidth]{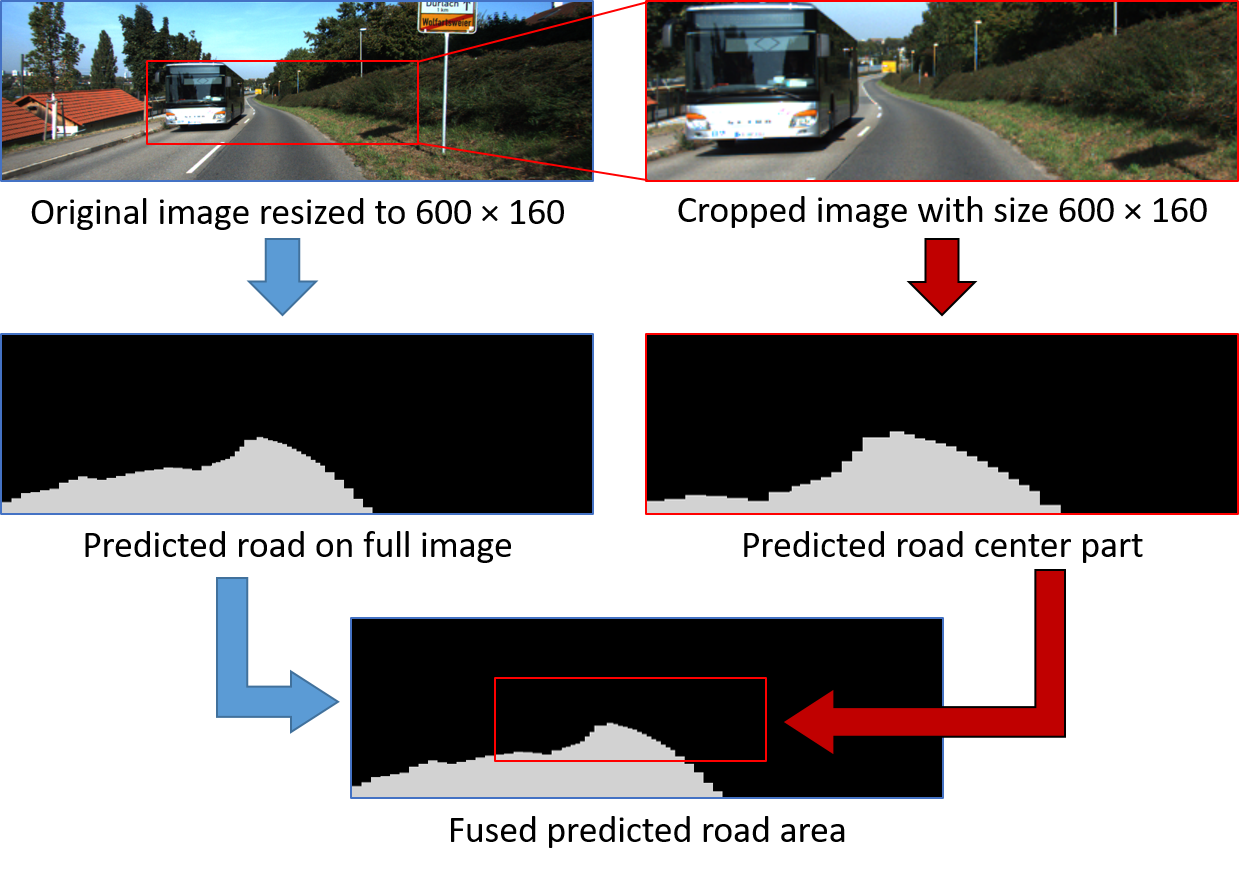}\caption{An illustration of the pyramid prediction scheme}
\label{Fig: Pyramid Prediction}
\end{figure}

After processing two input tensors in the network, we now have results
as two row vectors. The elements in the row vector denote the vertical
position of the drivable region boundary in the corresponding column.
To transform the row vectors to a drivable region on the camera frame,
we draw a polygon that connects corresponding vertices and the bottom
line. At last, we scale the first result map to the original size
and overwrite the center pixels using the second result map as illustrated
in Figure \ref{Fig: Pyramid Prediction}. 

\begin{table*}[!t]
\centering{}\caption{Comparison between networks on the KITTI\cite{fritsch2013KITTI_road}
road segmentation challenge: F1-score (F1), average precision (AP),
precision (PRE), recall (REC), false positive rate (FPR), false negative
rate (FNR), number of parameters (Para), floating operations (FL-OPs)
and runtime\label{Table: Evaluation result}}
\begin{tabular}{|c|c|c|c|c|c|c|c|c|c|}
\hline 
Method &
F1 &
AP &
PRE &
REC &
FPR &
FNR &
Para &
FLOPs &
Runtime\tabularnewline
\hline 
\hline 
ours &
89.08\% &
91.60\% &
88.12\% &
90.06\% &
6.69\% &
9.94\% &
0.35M &
6.9B &
16 ms\tabularnewline
\hline 
ours without LSTM &
81.84\% &
73.27\% &
81.66\% &
82.02\% &
10.15\% &
17.98\% &
0.36M &
7.0B &
16 ms\tabularnewline
\hline 
MAP\cite{laddha2016map} &
87.80 \% &
89.96\% &
86.01\% &
89.66\% &
8.04\% &
10.34\% &
457.43M &
7.15B &
280ms\tabularnewline
\hline 
StixelNet\cite{levi2015stixelnet} &
89.12\% &
81.23\% &
85.80\% &
92.71\% &
8.45\% &
7.29\% &
6.82M &
43.0B &
1s\tabularnewline
\hline 
Up-Conv-Poly\cite{oliveira2016UpConv} &
93.83\% &
90.47\% &
94.00\% &
93.67\% &
3.29\% &
6.33\% &
19.44M &
31.5B &
83ms\tabularnewline
\hline 
\end{tabular}
\end{table*}

\subsection{Network training and testing }

The proposed network is trained in KITTI road training set. There
are $289$ training samples and the image sizes range from $1242\times375$
to $1224\times370$. The ground truth of each sample is formatted
as a binary image with the same size. To augment the training set
we scale the sample images by $0.5$ and $1.0$ and then use a sliding
window of size $600\times160$ to capture the images and labels. The
stride of the sliding window is $60$ pixels in horizontal and $20$
pixels in vertical. Finally, $20,808$ samples are generated. Those
samples are separated into a $20,000$ sample training set and an
$808$ sample validating set. We also add Gaussian noise to the input
data with a standard deviation of $0.02\%$ for additional diversity.
Adam \cite{Kingma2015Adam} is a gradient descent based optimizer
that adjusts learning rate on each neuron according to the estimation
of lower-order moments of the gradients. We choose this optimizer
because it accelerates the converging process in the beginning and
slows it down near optimum. During training, the input batch size
is set to $100$ and reference learning rage is set to $1e^{-5}$.
After training $80$ epochs we get $0.0185$ mean average error on
validation data.

We then apply the network on the KITTI road testing set and submit
to the benchmark server. The result shows that the proposed solution
achieved $89.08\%$ in F1-score and $91.60\%$ in average precision
(AP), which are comparable to the existing approaches. In Table \ref{Table: Evaluation result}
our work is compared to the related solutions listed on the KITTI
road benchmark. It shows that our work has a similar F1-score and
AP to other works but a higher precision and lower false positive
rate. The proposed network has fewer parameters to train and takes
less floating-point operations to process. When testing on an NVidia
GTX 950M GPU, the proposed solution achieved 16 ms/frame network runtime.
We also trained a pure-CNN network by replacing the distributed LSTM
layers with $3\times3$ convolutional layers. The accuracy of the
pure-CNN solution is worse than that of CNN-LSTM solution, as shown
in Table \ref{Table: Evaluation result}.

\begin{figure}
\begin{centering}
\begin{tabular}{c}
\includegraphics[width=0.75\columnwidth]{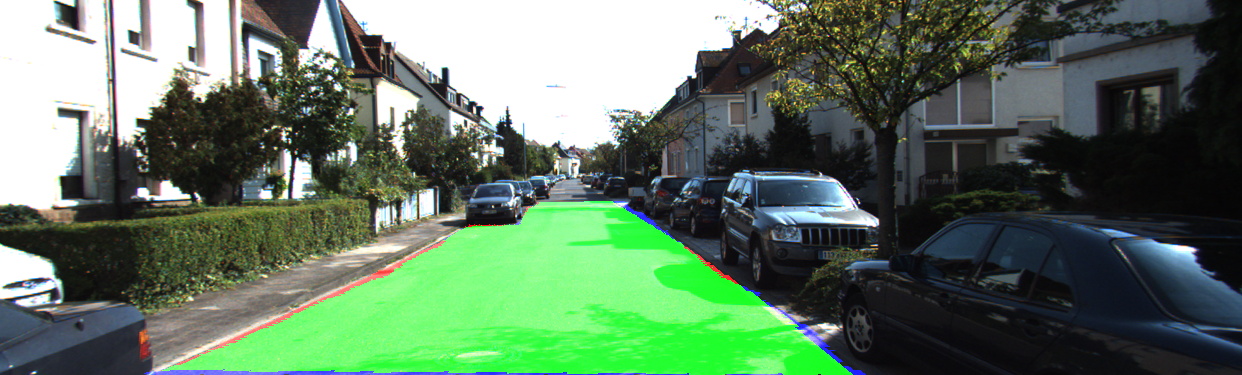}\tabularnewline
\includegraphics[width=0.75\columnwidth]{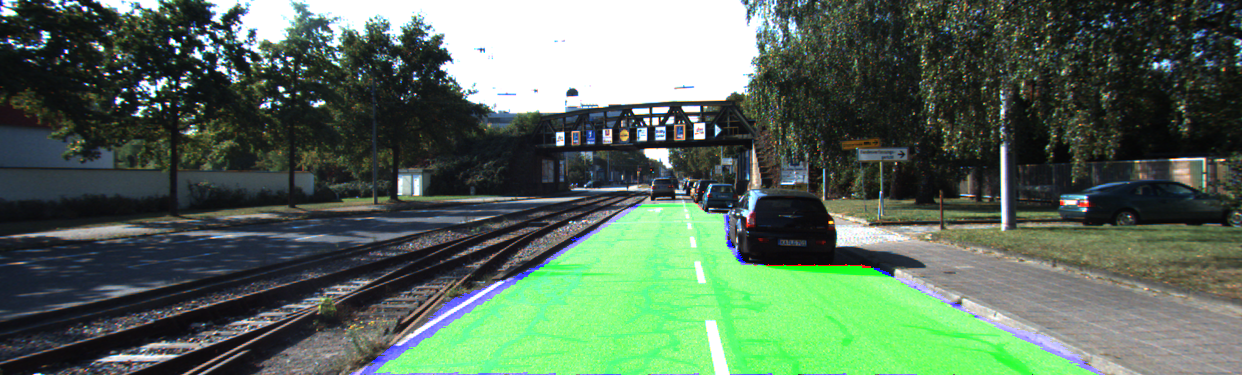}\tabularnewline
\end{tabular}
\par\end{centering}
\caption{Typical road segmentation results }

\label{Fig: Typical Result}
\end{figure}

Figure \ref{Fig: Typical Result} shows the typical result of the
proposed road detector. Green pixels are true positives, red pixes
are false negatives, and blue pixels are false positives. It is obvious
that most of the road surface is detected, and obstacles such as vehicles
and railways are separated to avoid collisions. False negatives occasionally
appear at road-vehicle and road-sidewalk interfaces , which is not
a safety concern for automated driving. However, false positive on
the sidewalks needs further corrections.

\section{Conclusion \label{sec:Conclusion}}

In this paper, we compare the convolutional layer and distributed
LSTM layer and demonstrate the advantages of combing the CNN and LSTM
structures for spatial feature map processing. We also propose a neural
network to evaluate its performance. The test result on the KITTI
road benchmark shows that our solution achieves 89.08\% in F1-score
and 91.60\% in average precision. However, the image-based road segmentation
is subjected to illumination conditions. Shadows, blurs, and ambiguous
textures can cause false positives and false negatives. In future
work, a fusion of multiple sensors including camera, LiDAR and radar
will be applied to improve the road segmentation as well as object
detection.

\section{Acknowledgement}

This work is supported by U.S. NSF Grant CNS-1626236.

\newpage{}

\bibliographystyle{unsrt}
\bibliography{5_home_yecheng_ISCAS2019_0Reference}

\end{document}